\documentclass[runningheads,a4paper]{llncs}

\usepackage{amssymb}
\usepackage{graphicx}
\usepackage{microtype}
\usepackage{subfigure}
\usepackage{amsmath}
\usepackage{multirow}
\usepackage{booktabs} % for professional tables
\usepackage{algorithm}
\usepackage{algorithmic}
\usepackage{float}
\usepackage{hyperref}

\usepackage{url}
\urldef{\mailsa}\path|{ziruiw, jgc}@cs.cmu.edu|

\begin{document}

\mainmatter  % start of an individual contribution

\title{Towards more Reliable Transfer Learning}


\author{Zirui Wang \and Jaime Carbonell}
%


\institute{
Language Technologies Institute\\
Carnegie Mellon University, Pittsburgh PA, USA\\
\mailsa}


\toctitle{Lecture Notes in Computer Science}
\tocauthor{Authors' Instructions}
\maketitle

\begin{abstract}
Multi-source transfer learning has been proven effective when within-target labeled data is scarce. Previous work focuses primarily on exploiting domain similarities and assumes that source domains are richly or at least comparably labeled. While this strong assumption is never true in practice, this paper relaxes it and addresses challenges related to sources with diverse labeling volume and diverse reliability. The first challenge is combining domain similarity and source reliability by proposing a new transfer learning method that utilizes both source-target similarities and inter-source relationships. The second challenge involves pool-based active learning where the oracle is only available in source domains, resulting in an integrated active transfer learning framework that incorporates distribution matching and uncertainty sampling. Extensive experiments on synthetic and two real-world datasets clearly demonstrate the superiority of our proposed methods over several baselines including state-of-the-art transfer learning methods.
\end{abstract}

\section{Introduction}

Traditional supervised machine learning methods share the common assumption that training data and test data are drawn from the same underlying distribution. Typically, they also require sufficient labeled training instances to construct accurate models. In practice, however, one often can only obtain limited labeled training instances. Inspired by human beings' ability to transfer previously learned knowledge to a related task, \textit{transfer learning} \cite{pan2010survey} addresses the challenge of data scarcity in the \textit{target} domain by utilizing labeled data from other related \textit{source} domain(s). Plenty of research has been done on the single-source setting \cite{raina2007self,pan2011domain,zhang2016lsdt}, and it has been shown that transfer learning has a wide range of applications in areas such as sentiment analysis \cite{blitzer2007biographies,pan2010cross}, computer vision \cite{long2017deep}, cross-lingual natural language processing \cite{moon2017completely}, and urban computing \cite{wei2016transfer}.

In recent years, new studies are contributing to a more realistic transfer learning setting with multiple source domains, both for classification problems \cite{yang2007cross,luo2008transfer,duan2009domain,sun2011two} and regression problems \cite{pardoe2010boosting,wei2017source}. These approaches try to capture the diverse \textit{source-target} similarities, either by measuring the distribution differences between each source and the target or by re-weighting source instances, so that only the source knowledge likely to be relevant to the target domain is transferred. However, these works assume that all sources are equally reliable; in other words, all sources have labeled data of the same or comparable quantity and quality. Nonetheless, in real-world applications, no such assumption holds. For instance, recent sources typically contain less labeled data than long-established ones, and narrower sources contain less data than broader ones. Also some sources may contain more labeling noise than others. Therefore, it is common for source tasks to exhibit diverse reliabilities. However, to the best of our knowledge, not much work has been reported in the literature to compensate for source reliability divergences.  
Ideally, we thrive to find sources that are relevant and reliable at the same time, as a compromise in either aspect can hurt the performance. When this is infeasible, one has to include some less reliable sources and carefully weigh the trade-off. There are two reasons why including a source that is \textbf{not} richly labeled is desired. First, such a source may be very informative since it is closely related to the target domain. For example, in low-resource language processing, a language that is very similar to the target language is usually also relatively low-resource but it would still be used as an important source due to its proximity. In addition, while labeled data may be scarce, unlabeled data is often easier to obtain and one can acquire labels for unlabeled data from domain experts, assuming a budget is available via \textit{active learning} \cite{settles2010active}. 

Active learning addresses the problem of data scarcity differently than transfer learning by querying new labels from an oracle for the most informative unlabeled data. A fruitful line of prior work has developed various active learning strategies \cite{nguyen2004active,konyushkova2017learning,murugesan2017active}. Recently, the idea of combining transfer learning and active learning has attracted increasing attention \cite{chattopadhyay2013joint,wang2014active,kale2015hierarchical}. Most of existing work assumes that there is plenty of labeled data in source domain(s) and one can query labels for the target domain from an oracle. However, this assumption is not always true. Since transfer learning is mainly applied in target domains with restrictions on access to data, such as sensitive private domains or novel domains that lack human knowledge to generate sufficient labeled data, often there is little or \textbf{no} readily-available oracle in the target domain. For instance, in the influenza diagnosis task, transfer learning can be applied to assist diagnosing a new flu strain (target domain) by exploiting historical data of related flu strains (source domains), but there may be little expertise regarding the new strain. It is, however, often possible to query labels for a selected set of unlabeled data in source domains, as these may represent previous better-studied flu strains. Task similarity, corresponding to flu-strain diagnosis or treatment in this instance, may be established via commonalities in symptoms exhibited by patients with the new or old flu strains, or via analysis of the various influenza genomes. To the best of our knowledge, active learning for multi-source transfer learning has not been deeply studied.

 In this paper, we focus on a novel research problem of transfer learning with multiple sources exhibiting diverse reliabilities and different relations to the target task. We study two related tasks. The first one is how to construct a robust model to effectively transfer knowledge from unevenly reliable sources, including those with less labeled data. The second task is how to apply active learning on multiple sources to stimulate better transfer, especially in such a scenario that sources have diverse quantities and qualities. Notice that these two tasks are related but can also be independent. For instance, in the low-resource language problem mentioned above, the first task is applicable but not the second one. This is because we may have an oracle for neither the source nor the target as they may both be of extremely limited resources. On the other hand, both tasks are relevant to the influenza example.
 
 For the first task, inspired by \cite{murugesan2017active}, we propose a \textit{peer-weighted multi-source transfer learning} method (\textit{PW-MSTL}) that jointly measures source proximity and reliability. The algorithm utilizes \textit{inter-source relationships}, which have been less studied previously, in two ways: (i) it associates the proximity coefficients based on distribution difference with the reliability coefficients measured using \textit{peers}\footnote{We define the peers of a source as the other sources weighted by inter-source relationships.} to learn their target model relevance, and (ii) when a source is not confident in the prediction for a testing instance, it is allowed to query its peers. By doing so, the proposed method allows knowledge transfer among sources and effectively utilizes label-scarce sources. For the second task, we propose an active learning framework, called \textit{adaptive multi-source active transfer} (\textit{AMSAT}), that builds on the idea of Kernel Mean Matching (KMM) \cite{huang2007correcting,gretton2007kernel} and uncertainty sampling \cite{tong2001support} to select unlabeled instances that are the most representative and avoid information redundancy. Experimental results, shown later, demonstrate that the combination of \textit{PW-MSTL} and \textit{AMSAT} significantly outperforms other competing methods. The proposed methods are generic and thus can be generalized to other learning tasks.

\section{Related Work}

\textbf{Transfer learning} : Transfer learning aims at utilizing source domain knowledge to construct a model with low generalization error in the target domain. Transfer learning with multiple sources is challenging because of distribution discrepancies and complex inter-source dependencies. Most existing work focuses on how to capture the diverse domain similarities. Due to stability and capability of measuring fine-grained similarities explicitly, ensemble approaches are widely used. In \cite{yang2007cross}, adaptive SVM (A-SVM) is proposed to adapt several trained source SVMs and learn model importance from meta-level features based on distributions. In contrast, \cite{duan2009domain} used Maximum Mean Discrepancy (MMD) \cite{gretton2007kernel} as source weights, while adding an additional manifold regularization based on a smoothness assumption of the target classifier. A more sophisticated two-stage weighting methodology based on distribution matching and conditional probability differences is presented in \cite{sun2011two}. However, the valuable unlabeled data is not utilized in their method. Moreover, these methods assume that all sources are equally reliable and they have the limitation that inter-source relationships are ignored.

\textbf{Multi-task learning}: Unlike multi-source transfer learning which focuses on finding a \textit{single} hypothesis for the target task, multi-task learning \cite{zhang2014regularization} tries to improve performance on \textit{all} tasks and finds a hypothesis for each task by adaptively leveraging related tasks. It is crucial to learn task relationships in multi-task learning problems, either used as priors  \cite{cavallanti2010linear} or learned adaptively \cite{murugesan2016adaptive}. While this is similar to the inter-source relationships we utilized in this paper, multi-task learning does not involve measuring the proximity between a source and the target nor the trade-off between proximity and reliability. 

\textbf{Active learning}: At the other end of the spectrum, active learning has been proven to be effective both empirically and theoretically. A new definition of sample complexity is proposed in \cite{balcan2010true} to show that active learning is strictly better than passive learning asymptotically. In recent years, there has been increasing interests in combining transfer learning and active learning to jointly solve the issue of insufficient labeled data. A popular representative of this family is the work by Chattopadhyay et al. \cite{chattopadhyay2013joint} where the JO-TAL method is proposed. It is a unified framework that jointly performs active learning and transfer learning by matching Hilbert space embeddings of distributions.  A key benefit of these methods is to address the cold start problem of active learning by introducing a version space prior learned from transfer learning \cite{yang2013theory}. However, unlike our approach, these methods all assume that source data are richly labeled and only perform active learning in the target domain.

\section{Problem Formulation}

For simplicity, we consider a binary classification problem for each domain, but the methods generalize to multi-class problems and are also applicable to regression tasks. We consider the following multi-source transfer learning setting that is common in real world when one tries to study a novel target domain with extremely limited resources. We denote by [N] consecutive integers
ranging from 1 to N. Suppose we are given $K$ auxiliary source domains, each contains both labeled and unlabeled data. Moreover, these sources typically have varying amount of labeled data (and thus with diverse reliabilities). Specifically, we denote by $S_k=S_k^L \cup S_k^U$ the dataset associated with the $k^{th}$ source domain. Here, $S_k^L=\{(x_i^{S_k^L}, y_i^{S_k^L})\}_{i=1}^{n_k^L}$ is the labeled set of size $n_k^L$, where $x_i^{S_k^L} \in \mathbb{R}^d$ is the $i^{th}$ labeled instance of the $k^{th}$ source and $y_i^{S_k^L}$ is its corresponding true label. Similarly, $S_k^U=\{x_i^{S_k^U}\}_{i=1}^{n_k^U}$ is the set of unlabeled data of size $n_k^U$. For the target domain, we assume that there is no labeled data available but we have sufficient unlabeled data, denoted as $T=\{x_i^{T}\}_{i=1}^{n_T}$ where instances are in the same feature space as sources. Furthermore, for the purpose of active learning, we assume that there is a uniform-cost oracle available only in the source domains, and the conditional probabilities $P(Y|X)$ are at least somewhat similar among both source and target domains.

\section{Proposed Approach}

\subsection{Motivation}
We first present the theoretical motivation for our methods. We analyze the expected target domain risk in the multi-source setting, making use of the theory of domain adaptation proved in \cite{ben2007analysis,blitzer2008learning}.

\textbf{Definition 1} {\itshape For a hypothesis space $\mathcal{H}$ for instance space $\mathcal{X}$, the symmetric difference hypothesis space $\mathcal{H}\Delta\mathcal{H}$ is the set of hypotheses defined as}
\begin{equation}\label{eq:1}
\mathcal{H}\Delta\mathcal{H} = \{ h(x)\oplus h'(x): h, h' \in \mathcal{H} \},
\end{equation}
{\itshape where $\oplus$ is the XOR function. Then the $\mathcal{H}\Delta\mathcal{H}$-divergence between any two distributions $D$ and $D'$ is defined as}
\begin{equation}\label{eq:2}
d_{\mathcal{H}\Delta\mathcal{H}}(D, D') \triangleq 2 \mathop{\sup}\limits_{A\in\mathcal{A}_{\mathcal{H}\Delta\mathcal{H}}}\Big|Pr_{D} [A]-Pr_{D'} [A]\Big|,
\end{equation}
{\itshape where $\mathcal{A}_{\mathcal{H}\Delta\mathcal{H}}$ is the measurable set of subsets of $\mathcal{X}$ that are the support of some hypothesis in $\mathcal{H}\Delta\mathcal{H}$.}

Now if we assume that all domains have the same amount of unlabeled data (which is a reasonable assumption since unlabeled data are usually cheap to obtain), i.e. $n_k^U=n_T=m'$, then we can show the following risk bound on the target domain.

\textbf{Theorem 1} {\itshape Let $\mathcal{H}$ be a hypothesis space of VC-dimension d and $\hat{d}_{\mathcal{H}\Delta\mathcal{H}}(S_i^U, T)$ be the empirical distributional distance between the $i^{th}$ source and the target domain, induced by the symmetric difference hypothesis space. Then, for any $\mu \in (0, 1)$ and any $\hat{h} = \sum_i \alpha_i \hat{h}_i$ where $\hat{h}_i \in \mathcal{H}$ and $\sum \alpha_i = 1$, the following holds with probability at least $1 - \delta$, }
\begin{equation}\label{eq:3}
\small
\begin{split}
 \epsilon_T(\hat{h}) & \leq \Bigg[\sum\limits_i^K \alpha_i \bigg[ \mu \Big[\hat{\epsilon}_{S_i}(\hat{h}_i) + \frac{1}{2} \hat{d}_{\mathcal{H}\Delta\mathcal{H}}(S_i^U, T)\Big] \\
 & + \frac{1-\mu}{K-1} \sum\limits_{j\neq i}^K \Big[\hat{\epsilon}_{S_j}(\hat{h}_i) + \frac{1}{2} \hat{d}_{\mathcal{H}\Delta\mathcal{H}}(S_j^U, T) \Big] \Bigg] \\  
& + \bigg(\sum\limits_i^K \alpha_i \sqrt{\frac{\mu^2}{\beta_i} + (\frac{1-\mu}{K-1})^2 \sum_{j\neq i}\frac{1}{\beta_j}}\bigg)\sqrt{\frac{d\log(2m)-\log(\delta)}{2m}} \bigg] \\
& + 4 \sqrt{\frac{2d \log(2m')+\log(\frac{4}{\delta})}{m'}} + \lambda_{\alpha, \mu},
\end{split}
\end{equation}
\normalsize 
 {\itshape where $\epsilon_{*}(h)$ is the expected risk of h in the corresponding domain, $m=\sum_i^K n_i^L$ is the sum of labeled sizes in all sources, $\beta_i = n_i^L/m$ is the ratio of labeled data in the $i^{th}$ source, and $\lambda_{\alpha, \mu}$ is the risk of the ideal multi-source hypothesis weighted by $\alpha$ and $\mu$.}

The proof can be found in the supplementary material. By introducing a concentration factor $\mu$, we replace the size of labeled data for each source $n_i^L$ with the total size $m$ in the third line in Eq.(\ref{eq:3}), resulting in a tighter bound. Suppose that the hypothesis $\hat{h}_i$ is learned using data in the $i^{th}$ source, then the bound suggests to use peers to evaluate the reliability of $\hat{h}_i$ while the optimal weights should consider both proximity and reliability of the $i^{th}$ source. This inspired us to propose the following method.

\subsection{Peer-weighted Multi-source Transfer Learning}

\begin{algorithm}[tbh]
   \caption{PW-MSTL}
   \label{alg:1}
\begin{algorithmic}[1]
   \STATE {\bfseries Input:} $S=S^L \cup S^U$: source data; $T$: target data; $\mu$: concentration factor; $b_1$: confidence tolerance; $T$: test data size; 
   \FOR{$k=1,...,K$}
   \STATE Compute $\alpha^k$ by solving (6).
   \STATE Train a classifier $\hat{h}_k$ on the $\alpha^k$ weighted $S_k^L$.
   \ENDFOR
   \STATE Compute $\delta$ and $\mathbf{R}$ as explained in Section 4.2.
   \STATE Compute $\omega$ as (5).
   \FOR{$t=1,...,T$}
   \STATE Observe testing example $x^{(t)}$.
   \FOR{$k=1,...,K$}
   \IF{$|\hat{h}_k(x^{(t)}| < b_1$}
   \STATE Compute $\hat{p}_k^{(t)} = \sum\limits_{m \in [K], m \neq k} \mathbf{R}_{km}|\hat{h}_m(x^{(t)})|$.
   \ELSE
   \STATE Compute $\hat{p}_k^{(t)} = |\hat{h}_k(x^{(t)})|$.
   \ENDIF
   \ENDFOR
   \STATE Predict $\hat{y}^{(t)} = sign(\sum_{k\in [K]} \omega_k\hat{p}_k^{(t)})$.
   \ENDFOR
\end{algorithmic}
\end{algorithm}

In this paper, we propose the idea of formulating all source domains jointly in a framework similar to the multi-task learning. Similar to the task relationships in multi-task learning, we learn the inter-source relationships for the multi-source transfer learning problem by training a source relationship matrix $\mathbf{R} \in \mathbb{R}^{K \times K}$ as follows:
\begin{equation}\label{eq:4}
 \mathbf{R}_{i,j} = 
  \begin{cases} 
   \frac{\exp(\beta_1 \hat{\epsilon}_{S_i}(\hat{h}_j))}{\sum\limits_{j' \in [K], j' \neq i} \exp(\beta_1 \hat{\epsilon}_{S_i}(\hat{h}_j'))}, & i \neq j \\
   0,       & \text{otherwise}
  \end{cases}
\end{equation}
where $\hat{h}_j$ is the classifier trained on the $j^{th}$ source, $\hat{\epsilon}_{S_i}(\hat{h}_j)$ is the empirical error of $\hat{h}_j$ measured on the $i^{th}$ source and $\beta_1$ is a parameter to control the spread of the error. In other words, $\mathbf{R}$ is a matrix such that all entries on the diagonal are 0 and the $k^{th}$ row, namely $\mathbf{R}_k \in \Delta^{K-1}$, is a distribution over all sources where $\Delta^{K-1}$ denotes the probability simplex. Note that we do not require $\mathbf{R}$ to be symmetric due to the asymmetric nature of information transferability. While classifiers trained on a more reliable source can be well transferred to a less reliable source, the reverse is not guaranteed to be true.

A key benefit of the matrix $\mathbf{R}$ is that it measures source reliabilities directly. The bound in Eq.(\ref{eq:3}) suggests that this direct measurement of reliability gives the algorithm extra confidence in lower generalization error. Intuitively, if a classifier $\hat{h}_i$ trained on the $i^{th}$ source has a low empirical error on the $j^{th}$ source and the distributional divergence (such as the $\mathcal{H}\Delta\mathcal{H}$-divergence) between the $j^{th}$ source and the target is small, then $\hat{h}_i$ should have a low error on the target domain as well. This is shown by the second line in Eq.(\ref{eq:3}).

We therefore parametrize the source importance weights, considering both source proximity and reliability, as follows:
\begin{equation}\label{eq:5}
\omega = \delta \cdot [\mu\mathbf{I_{K}} + (1-\mu)\mathbf{R}]
\end{equation}
where $\mathbf{I_{K}} \in \mathbb{R}^{K \times K}$ is the identity matrix, $\delta \in \mathbb{R}^{K} $ is a vector measuring pairwise source-target proximities, $\mu \in [0,1]$ is a scalar and $\mathbf{R}$ is the source relationship matrix. The concentration factor $\mu$ introduces an additional degree of freedom in quantifying the trade-off between proximity and reliability. Setting $\mu = 1$ amounts to weight sources based on proximity only. In the next section, we obtain the effective heuristic of specifying $\mu=\frac{1}{K}$ in our experiments. The proximity vector $\delta$ may be manually specified according to domain knowledge or estimated from data. In this paper, we adapt the Maximum Mean Discrepancy (MMD) statistic \cite{huang2007correcting,gretton2007kernel} to estimate the distribution discrepancy. It measures the difference between the means of two distributions after mapping onto a Reproducing Kernel Hilbert Space (RKHS), avoiding the complex density estimation. Specifically, for the $k^{th}$ source, its re-weighted MMD can be written as:
\begin{equation}\label{eq:6}
\mathop{\min}\limits_{\alpha^k} \Big\Vert \frac{1}{n_k^L+n_k^U} \sum \limits_{i=1}^{n_k^L+n_k^U} \alpha_i^k \Phi(x_i^{S_k}) - \frac{1}{n_T} \sum \limits_{i=1}^{n_T} \Phi(x_i^{T}) \Big\Vert_H^2
\end{equation} 
where $\alpha_i^k \geq 0$ are the weights of the $k^{th}$ source aggregate data and $\Phi(x)$ is a feature map onto the RKHS $H$. By applying the kernel trick, the optimization in (6) can be solved efficiently as a quadratic programming problem using interior point methods.

For label-scarce sources, the ensemble approach often fails to utilize their information effectively. This unsatisfactory performance leads us to propose using confidence and source relationships to transfer knowledge. For many standard classification methods such as SVMs and perceptrons, we can use the distance to the boundary $|\hat{h}_k(x)|$ to measure the confidence of the $k^{th}$ classifier on the example $x$, as in previous works \cite{murugesan2017active}. If the confidence is low, we allow the classifier to query its peers on this specific example, exploiting the source relationship matrix $\mathbf{R}$. Algorithm \ref{alg:1} summarizes our proposed method.

\subsection{Adaptive Multi-source Active Learning}

When an oracle is available in the source domain, we can acquire more labels for source data, especially for less reliable but target-relevant sources. This is different from traditional active learning, which tries to improve the accuracy within the target domain. The TLAS algorithm proposed in \cite{huang2016transfer} performs the active learning on the source domain in the single-source transfer learning, by solving a biconvex optimization problem. However, their solution requires solving two quadratic programming problems alternatively at each iteration and is therefore computationally expensive, especially for multi-source problems of large scale. In this paper, we propose an efficient two-stage active learning framework. The pseudo-code is in Algorithm \ref{alg:2}.

In the first stage, the algorithm selects the source domain for query in an adaptive manner. The idea is that while we want to explore label-scarce sources for high marginal gain in performance, we also want to exploit label-rich sources for reliable queries. Eq.(\ref{eq:3}) suggests that the learner should prefer a uniform source labeled ratio. Therefore, we draw a Bernoulli random variable $P^{(t)}$ with probability $D_{KL}(\beta || uniform)$, i.e. the Kullback-Leibler (KL) divergence between the current ratio of source labeled data and the uniform distribution. If $P^{(t)} = 1$, then the algorithm explores sources that contain less labeled data. If $P^{(t)} = 0$, the algorithm exploits sources as demonstrated in line 12 of Algorithm \ref{alg:2}. Notice that the combined weights proposed in Eq.(\ref{eq:5}) play a crucial role here by providing a measurement of sources with unequal reliabilities.

\begin{algorithm}[tb]
   \caption{AMSAT}
   \label{alg:2}
\begin{algorithmic}[1]
   \STATE {\bfseries Input:} $S=S^L \cup S^U$: source data; $T$: target data; $\mu$: concentration factor; $B$: budget; 
   \FOR{$k=1,...,K$}
   \STATE Compute $\alpha^k$ by solving (6).
   \STATE Train a classifier $\hat{h}_k$ on the $\alpha^k$ weighted $S_k^L$.
   \ENDFOR
   \FOR{$t=1,...,B$}
   \STATE Compute $\beta_i^{(t)} = \frac{n_i^L}{\sum_i n_i^L}$. 
   \STATE Draw a Bernoulli random variable $P^{(t)}$ with probability $D_{KL}(\beta^{(t)} || uniform)$.
   \IF{$P^{(t)} = 1$}
   \STATE Set $Q^{(t)} = \frac{1}{\beta^{(t)}}$.
   \ELSE
   \STATE Compute $\omega^{(t)}$ as (5) and set $Q^{(t)} = \omega^{(t)}$.
   \ENDIF
   \STATE Draw $k^{(t)}$ from [K] with distribution $Q^{(t)}$. 
   \STATE Select $x^{(t)}$ according to (8) and query the label for it.
   \STATE Update $S_{k^{(t)}}^{L} \leftarrow S_{k^{(t)}}^{L} \cup \{x^{(t)}\}$.
   \STATE Update $S_{k^{(t)}}^{U} \leftarrow S_{k^{(t)}}^{U} \setminus \{x^{(t)}\}$.
   \STATE Update classifier $\hat{h}_{k^{(t)}}$.
   \ENDFOR

\end{algorithmic}
\end{algorithm}

In the second stage, the algorithm queries the most informative instance within the selected source domain. Nguyen and Smeulders \cite{nguyen2004active} propose a Density Weighted Uncertainty Sampling criterion as:
\begin{equation}\label{eq:7}
 x = \mathop{\arg\max}\limits_{x_i \in S_{k^{(t)}}^U} E[(\hat{y}_i - y_i)^2| x_i]p(x_i)
\end{equation}
which picks the instance that is close to the boundary and relies in a denser neighborhood. In our setting, we propose to combine the distribution matching weights $\alpha$ in Eq.(\ref{eq:6}) and the uncertainty sampling to form the following selection criterion:
\begin{equation}\label{eq:8}
 x = \mathop{\arg\max}\limits_{x_i \in S_{k^{(t)}}^U} E[(\hat{y}_i - y_i)^2| x_i]\alpha^{k^{(t)}}_{i}
\end{equation}
This criterion selects instances that are representative in both source and target domains while it also takes uncertainty into consideration. Notice that we solve Eq.(\ref{eq:6}) once only and store the value of $\alpha$. As shown in \cite{settles2008analysis}, such approach is as efficient as the base informativeness measure, i.e. uncertainty sampling. Therefore, the proposed method achieves efficiency, representativeness and minimum information overlap.

\section{Empirical Evaluation}

In the following experimental study we aim at two goals: (i) to evaluate the performance of \textit{PW-MSTL} with sources with diverse reliabilities, and (ii) to evaluate the effectiveness of \textit{AMSAT} in constructing more reliable classifiers via active learning. Due to space limitation, not all results are presented, but they show similar patterns. Unless otherwise specified, all model parameters are chosen via 5-fold cross validation and all results are averaged over experiments repeated randomly 30 times.

\subsection{Datasets}

\textbf{Synthetic dataset.} We generate a synthetic data for 5 source domains and 1 target domain. The samples $x \in \mathbb{R}^{10}$ are drawn from Gaussian distributions $\mathcal{N}(\mu_T+\mathbf{p}\Delta\mu, \sigma)$, where $\mu_T$ is the mean of the target domain, $\Delta\mu$ is a random fluctuation vector and $\mathbf{p}$ is the variable controlling the proximity between each source and the target (higher $\mathbf{p}$ indicates lower proximity). We then consider a labeling function $f(x) = sign((w_0^T + \delta\Delta w)x + \epsilon)$, where $w_0$ is a fixed base vector, $\Delta w$ is a random fluctuation vector and $\epsilon$ is a zero-mean Gaussian noise term. We set $\delta$ to small values as we assume labeling functions are similar. Using different $\mathbf{p}$ values, We generate 50 positive points and 50 negative points as training data for each source domain and additional 100 balanced testing samples for the target domain.

\textbf{Spam Detection.}\footnote{\url{http://ecmlpkdd2006.org/challenge.html}} We use the task B challenge of the spam detection dataset from ECML PAKDD 2006 Discovery challenge. It consists of data from inboxes of 15 users and each user forms a single domain with 200 spam ($+$) examples and 200 non-spam ($-$) examples. Each example consists of approximately 150$K$ features representing word frequencies and we reduce the dimension to 200 using the latent semantic
analysis (LSA) method \cite{halko2011finding}. Since some spam types are shared among users while others are not, these domains form a multi-source transfer learning problem if we try to utilize data from other users to build a personalized spam filter for a target user.

\textbf{Sentiment Analysis.}\footnote{\url{http://www.cs.jhu.edu/~mdredze/datasets/sentiment}} The dataset contains product reviews from 24 domains on Amazon. We consider each domain as a binary classification task by setting reviews with rating $>3$ as positive ($+$) and reviews with rating $<3$ as negative ($-$), while reviews with rating $=3$ were discarded as they are ambiguous. We pick the 10 domains that each contains more than 2000 samples and for each round of experiment, we randomly draw 1000 positive reviews and 1000 negative reviews for each domain. Similar to the previous dataset, each review contains of approximately 350$K$ features and we reduce the dimension to 200 using the same method.

For each set of experiments, we have 600 examples (100 examples per domain) for \textit{synthetic}, 6000 emails (400 emails per user) for \textit{spam} and 20000 reviews (2000 reviews per domain) for \textit{sentiment}. We set one domain as the target and the rest as sources. For each source domain, we randomly divide the data into the labeled set and the unlabeled set, using a labeled-data fraction. For the target domain, we randomly divide it into two parts: $40\%$ for testing and $60\%$ as unlabeled training data (with the exception of \textit{synthetic} as we have generated extra testing data). Note that we can set different labeled-data fractions for source domains to model diverse reliabilities.

\subsection{Reliable Multi-source Transfer Learning}

\textbf{Competing Methods.} To evaluate the performance of our approach, we compare the proposed PW-MSTL with five different methods: one single-source method, one aggregate method, and three ensemble methods. Specifically, Kernel Mean Matching (KMM) \cite{huang2007correcting} is a conventional single-source method by distribution matching. We perform KMM for each single source and report the best performance (note that this is impractical in general as it requires an oracle to know which classifier is the best). Kernel Mean Matching-Aggregate (KMM-A) is the aggregate method which performs KMM on all sources' training data combined. For the ensemble approach, Adaptive SVM (A-SVM)
\cite{yang2007cross} and Domain Adaptation Machine (DAM) \cite{duan2009domain} are two widely-adopted multi-source methods. Finally, we also compare with our own baseline $\text{PW-MSTL}_{b}$, which is similar to DAM but directly uses Eq.(\ref{eq:5}) as model weights. We also compared with Transfer Component Analysis (TCA) \cite{pan2011domain} but the result is omitted due to its similar performance to KMM.

\begin{figure*}[tbh]
\begin{center}
  \subfigure[Accuracy on \textbf{dvd}]{ 
    \includegraphics[width=.31\columnwidth]{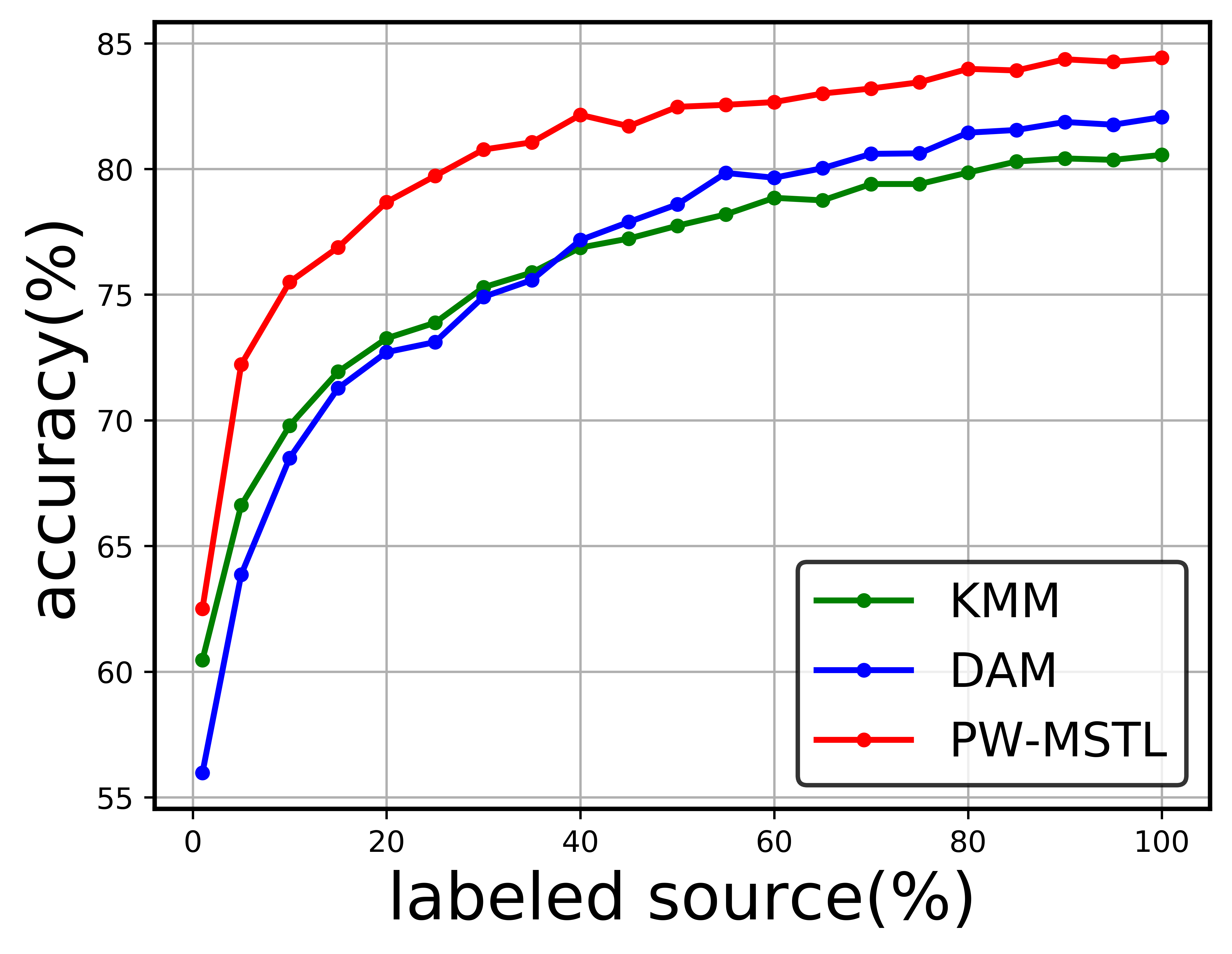} \label{fig:1a} 
  } 
  \hfill 
  \subfigure[Sensitivity of $\mu$ (uneven sources)]{% 
    \includegraphics[width=.31\columnwidth]{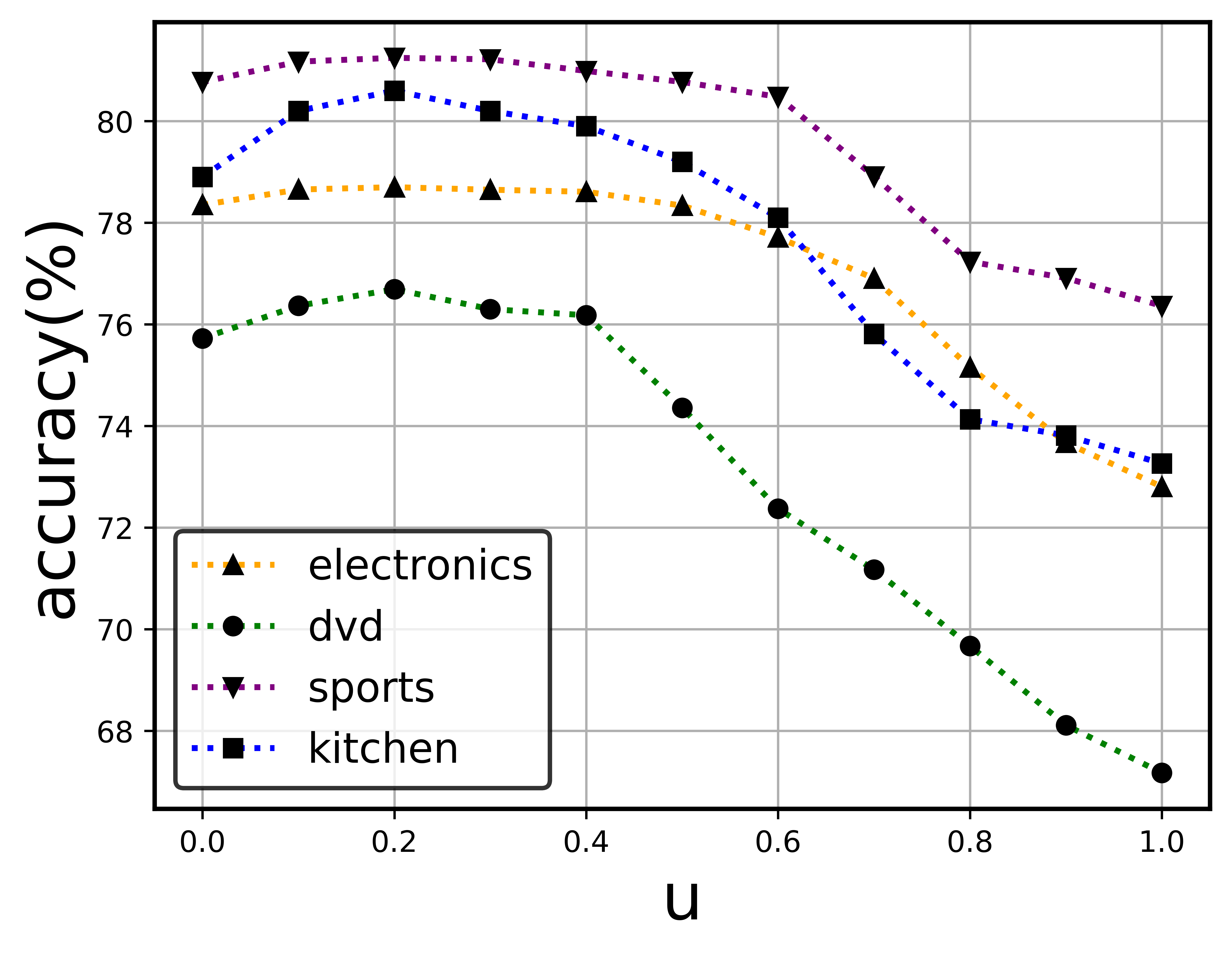} \label{fig:1b} 
  }
  \hfill 
  \subfigure[Sensitivity of $\mu$ (50\% labeled sources)]{% 
    \includegraphics[width=.31\columnwidth]{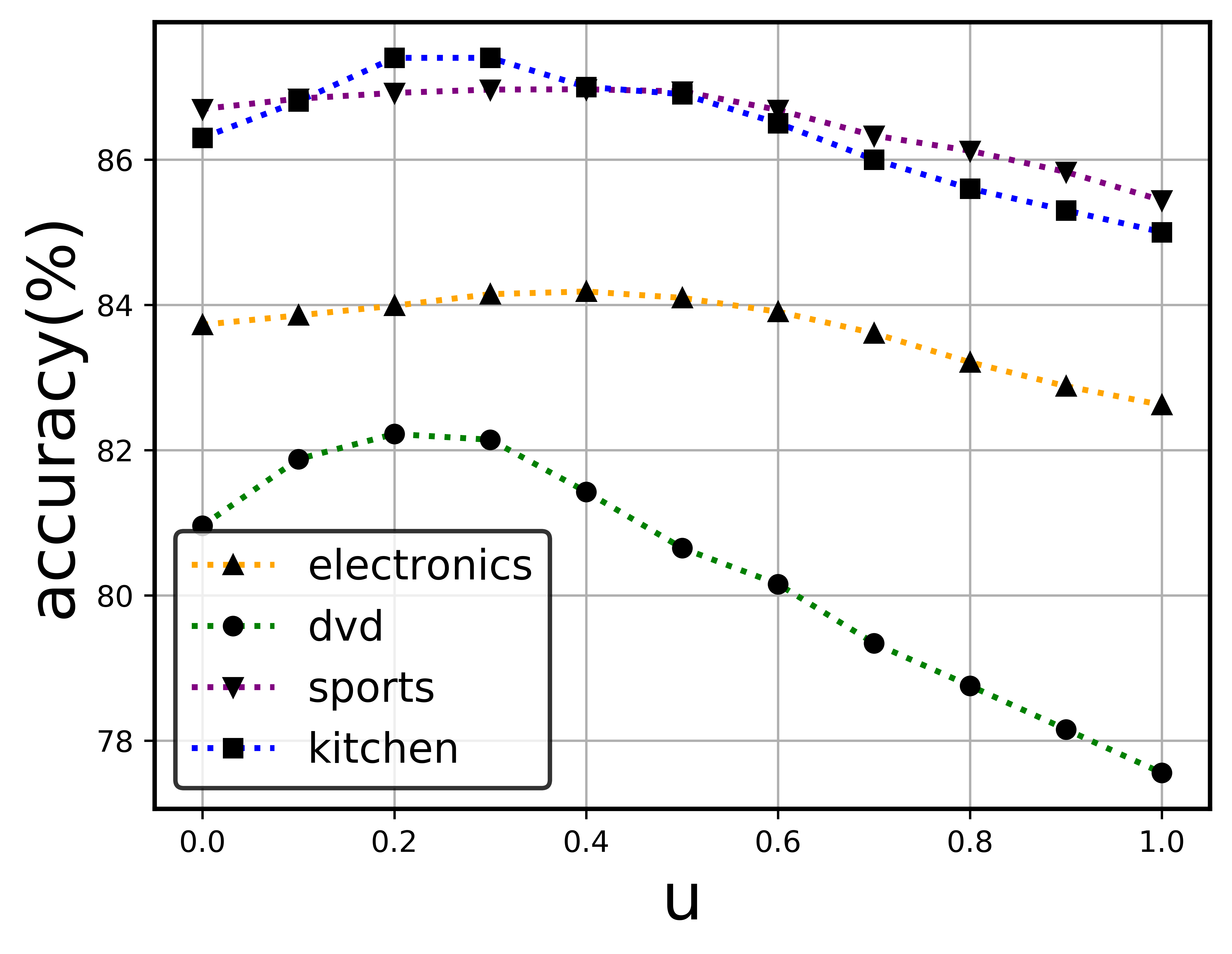} \label{fig:1c} 
  } 
\caption{Empirical analysis: (a) incremental accuracy on \textbf{dvd} of KMM, DAM \& PW-MSTL, when sources have same amount of labeled data; (b) sensitivity of $\mu$ when sources have different amount of labeled data; (c) sensitivity of $\mu$ when sources all have 50\% of labeled data.}
\label{fig:1}
\end{center}
\end{figure*}

\textbf{Setup.} For all methods, we use SVM as the base model and fix $b_1 = 1$. For competing methods, we mainly follow the standard procedures for model selection as explained in their respective papers. For fair comparisons, we use the same Gaussion kernel $k(x_i, x_j)=e^{-\Vert x_i-x_j\Vert^2/\gamma}$ with the bandwidth $\gamma$ set to the median pairwise distance of training data according to the \textit{median heuristic} \cite{gretton2012optimal}. Using other kernels yields similar results. For ensemble methods, we set the proximity weight as $\delta_k = \frac{exp(-\beta_2 MMD^{\rho}(S_k, T))}{\sum\limits_k exp(-\beta_2 MMD^{\rho}(S_k, T))}$, where $\beta_2$ and $\rho$ are tuned for each dataset to control the spread of MMD as described in Section 4.2. For both PW-MSTL and $\text{PW-MSTL}_{b}$, we set $\mu=0.2$ for all experiments and use the 0-1 loss to compute the relation matrix $\mathbf{R}$ (See Figure \ref{fig:1} for results on varying $\mu$).

\begin{table*}[t]
\caption{Classification accuracy ($\%$) on the target domain, given that source domains contain diverse \{1\%,5\%,15\%,30\%\} labeled data.}
\label{table1}
\vskip 0.1in
\begin{center}
\begin{tiny}
\begin{tabular}{l c c c c c c c c c c c c c c c}
\toprule
\multirow{2}{*}{Method} & \multicolumn{2}{c}{Synthetic} & \multicolumn{3}{c}{Spam} & \multicolumn{10}{c}{Sentiment} \\
\cmidrule(lr){2-3} \cmidrule(lr){4-6} \cmidrule(lr){7-16}
& case1 &  case2 & user7 & user8 & user3 & electronics & toys & music & apparel & dvd & kitchen & video & sports & book & health \\
\midrule
KMM   & 82.7 & 88.8 & 92.0 & 91.8 & 89.7 & 77.6 & 77.4 & 71.0 & 78.3 & 72.4 & 78.4 & 72.1 & 79.1 & 71.2 & 77.4\\
KMM-A & 87.3 & 91.4 & 92.0 & 92.0 & 91.8 & 74.6 & 76.3 & 70.3 & 75.8 & 72.4 & 75.2 & 70.5 & 76.7 & 69.7 & 74.9 \\
A-SVM & 70.8 & 89.4 & 84.5 & 87.8 & 86.8 & 70.8 & 73.7 & 67.7 & 73.6 & 62.6 & 72.8 & 62.5 & 73.7 & 66.9 & 71.4 \\
DAM    & 75.8 & 91.0 & 83.8 & 85.4 & 86.8 & 71.3 & 73.7 & 68.0 & 75.1 & 62.5 & 72.1 & 62.0 & 73.0 & 68.0 & 72.5 \\
$\text{PW-MSTL}_{b}$ & 85.5 & 90.8 & 91.5 & 92.6 & 90.3 & 78.0 & 78.7 & 70.7 & 79,5 & 73.2 & 78.3 & 72.5 & 79.5 & 71.5 & 77.7\\
PW-MSTL & \underline{\bf 88.4} & \underline{\bf 92.6} & \underline{\bf 93.8} & \underline{\bf 95.6} & \underline{\bf 92.8} & \underline{\bf 79.3} & \underline{\bf 81.9} & \underline{\bf 74.6} & \underline{\bf 82.7} &\underline{ \bf 76.7} & \underline{\bf 80.7} & \underline{\bf 76.2} & \underline{\bf 82.7} & \underline{\bf 74.8} & \underline{\bf 80.9} \\
\bottomrule
\end{tabular}
\end{tiny}
\end{center}
\vskip -0.1in
\end{table*}

\begin{table*}[t]
\caption{Classification accuracy ($\%$) on the target domain, given that source domains contain the same fraction ($\%_{L}$) of labeled data.}
\label{table2}
\vskip 0.1in
\begin{center}
\begin{tiny}
\begin{tabular}{c l c c c c c c c c c c c c c c}
\toprule
 \multirow{2}{*}{$\%_{L}$} & \multirow{2}{*}{Method} & \multirow{2}{*}{Synthetic} & \multicolumn{3}{c}{Spam} & \multicolumn{10}{c}{Sentiment} \\
\cmidrule(lr){4-6} \cmidrule(lr){7-16}
& & & user7 & user8 & user3 & electronics & toys & music & apparel & dvd & kitchen & video & sports & book & health \\
\midrule
\multirow{6}{*}{10$\%$} & KMM   & 87.0 & 89.1 & 91.2 & 90.3 & 75.0 & 74.6 & 68.3 & 75.6 & 70.2 & 75.9 & 69.9 & 75.6 & 68.9 & 74.3\\
& KMM-A & 91.1 & 91.3 & 90.7 & 91.0 & 74.8 & 76.5 & 70.2 & 76.8 & 71.3 & 77.6 & 71.6 & 77.7 & 71.3 & 75.4 \\
& A-SVM  & 89.4 & 88.4 & 91.9 & 89.2 & 77.1 & 78.1 & 69.9 & 78.2 & 68.9 & 79.1 & 69.2 & 78.1 & 70.5 & 77.1 \\
& DAM     & 89.7 & 89.6 & 90.4 & 91.3 & 77.5 & 79.0 & 69.9 & 79.8 & 69.0 & 79.5 & 68.9 & 78.4 & 71.9 & 77.7 \\
& $\text{PW-MSTL}_{b}$  & 90.2 & 89.7 & 92.4 & 92.1 & 77.7 & 78.7 & 69.7 & 78.9 & 73.5 & 79.8 & 72.3 & 78.8 & 70.4 & 77.9\\
& PW-MSTL  & 91.2 & \underline{\bf 92.5} &  \underline{\bf 94.9} & \underline{\bf 93.1} & \underline{\bf 79.8} & \underline{\bf 81.5} & \underline{\bf 73.3} & \underline{\bf 81.3} & \underline{\bf 76.4} & \underline{\bf 82.3} & \underline{\bf 75.4} & \underline{\bf 81.2} & \underline{\bf 74.4} & \underline{\bf 80.7} \\
\midrule
\multirow{6}{*}{50$\%$} & KMM   & 95.6 & 92.6 & 94.0 & 91.8 & 81.6 & 81.7 & 75.0 & 82.2 & 76.9 & 83.0 & 77.5 & 82.8 & 75.3 & 81.2\\
& KMM-A & 97.2 & 91.4 & 93.8 & \underline{\bf 94.7} & 80.4 & 82.4 & 74.5 & 82.7 & 77.1 & 83.8 & 76.5 & 82.8 & 76.0 & 79.6 \\
& A-SVM  & 96.4 & 91.5 & 95.2 & 93.4 & 81.7 & 83.4 & 74.7 & 84.3 & 76.0 & 85.4 & 75.3 & 83.3 & 76.0 & 82.1 \\
& DAM     & 96.6 & 92.7 & 93.1 & 93.2 & 83.5 & 84.5 & 73.4 & 84.4 & 77.3 & 86.7 & 76.5 & 84.8 & 76.8 & 83.6 \\
& $\text{PW-MSTL}_{b}$  & 96.6 & 92.9 & 95.2 & 93.5 & 83.6 & 84.7 & 74.4 & 85.0 & 80.4 & 85.9 & 79.4 & 85.7 & 77.0 & 84.1\\
& PW-MSTL  & 97.2 & \underline{\bf 94.5} & \underline{\bf 95.7} & 93.7 & \underline{\bf 84.8} & \underline{\bf 86.4} & \underline{\bf 76.9} & \underline{\bf 87.2} & \underline{\bf 82.0} & \underline{\bf 87.6} & \underline{\bf 81.3} & \underline{\bf 87.3} & \underline{\bf 79.8} & \underline{\bf 86.4} \\
\bottomrule
\end{tabular}
\end{tiny}
\end{center}
\vskip -0.1in
\end{table*}

\textbf{Results and Discussion.} Table \ref{table1}\footnote{Due to space limits, we randomly show results of only three cases for \textit{Spam} since their overall results are similar.} shows the classification accuracy of different methods on various datasets when sources contain different labeled fractions randomly chosen from the set $\{1\%,5\%,15\%,30\%\}$, while Table \ref{table2} shows the results when sources have the same amount of labeled data. The method that outperforms all other methods within the column with $p<0.001$ significance is highlighted and underlined (and thus we omit standard deviation due to space). We can observe that our proposed PW-MSTL method outperforms other methods in most cases. When source reliabilities are uneven, we get a significant improvement in the test set accuracy. By comparing results across the two tables, we can also observe that gap between our methods and other methods diminishes as source domains acquire more labeled data (more reliable).

Note that we have built two scenarios for the \textit{Synthetic} dataset in Table \ref{table1}. We control the proximity variable $\mathbf{p}$ and the labeled fraction such that similar sources also contain more labeled data in case 1 while this is reversed in case 2. As a result, we observe that prior methods focusing on proximity measure obtain an accuracy boost under case 2 compared to case 1 while our methods are consistent. We also note that A-SVM and DAM perform poorly on \textbf{dvd} and \textbf{video} even when sources are balanced. This is because the distribution divergence between these two domains are very small, causing prior methods failure to utilize other less related sources effectively. We plot an incremental performance comparison on \textbf{dvd} in Figure \ref{fig:1a}, showing that DAM slowly catches up with the single-source method as labeled fractions increase.

Finally, we study the sensitivity of $\mu$. Figure \ref{fig:1b} and \ref{fig:1c} show the variation in accuracy for some cases in Table \ref{table1} and \ref{table2} with varying $\mu$ over [0,1]. We can observe that similar patterns that accuracy values first increase and then decrease as $\mu$ increases in both cases, only that the drop in the performance is smoother in Figure \ref{fig:1c}. This confirms the motivation of combining proximity and reliability and the theoretical result established in this paper. Results on other datasets are similar but we omit them in the graph due to different scales.

\subsection{Multi-source Active Learning}

\textbf{Baselines.} To the best of our knowledge, there is no existing study that can be directly applied to our setting. Therefore, to evaluate the performance of our proposed AMSAT algorithm, we compare to the following standard baselines:\\
(1)\textbf{Random}: Selects the examples to query randomly from all source domains. This is equivalent to passive learning.\\
(2)\textbf{Uncertainty}: This is a popular active learning strategy called uncertainty sampling, which picks the examples that are most uncertain or close to the decision boundary.\\
(3)\textbf{Representative}: Another widely used method that chooses the examples that are most representative according to distribution matching in Eq.(\ref{eq:6}).\\
(4)\textbf{Proximity}: Selects the source examples that are most similar to the target domain. Note that this method usually queries examples in the very few most similar sources.

\textbf{Setup.} Unlike traditional active learning where domains for active selection are label-scarce, we initialize our source domains with various amount of labeled data such that some sources are more richly labeled. For each set of experiments, we initialize each source domain with a labeled fraction randomly chosen from a fixed set. This same data partition is used for all comparing methods. After each query, we perform a multi-source transfer learning algorithm and record the accuracy on the test target data. The data partition is repeated randomly for 30 times and the average results are reported. For all of our experiments, we set the query budget to 10$\%$ of the total number of source examples in the dataset. For fair comparison, the selected examples are evaluated using the same transfer learning method.

\begin{figure*}[tbh]
\begin{center}
  \subfigure[Toys]{ 
    \includegraphics[width=.23\columnwidth, height=.23\columnwidth]{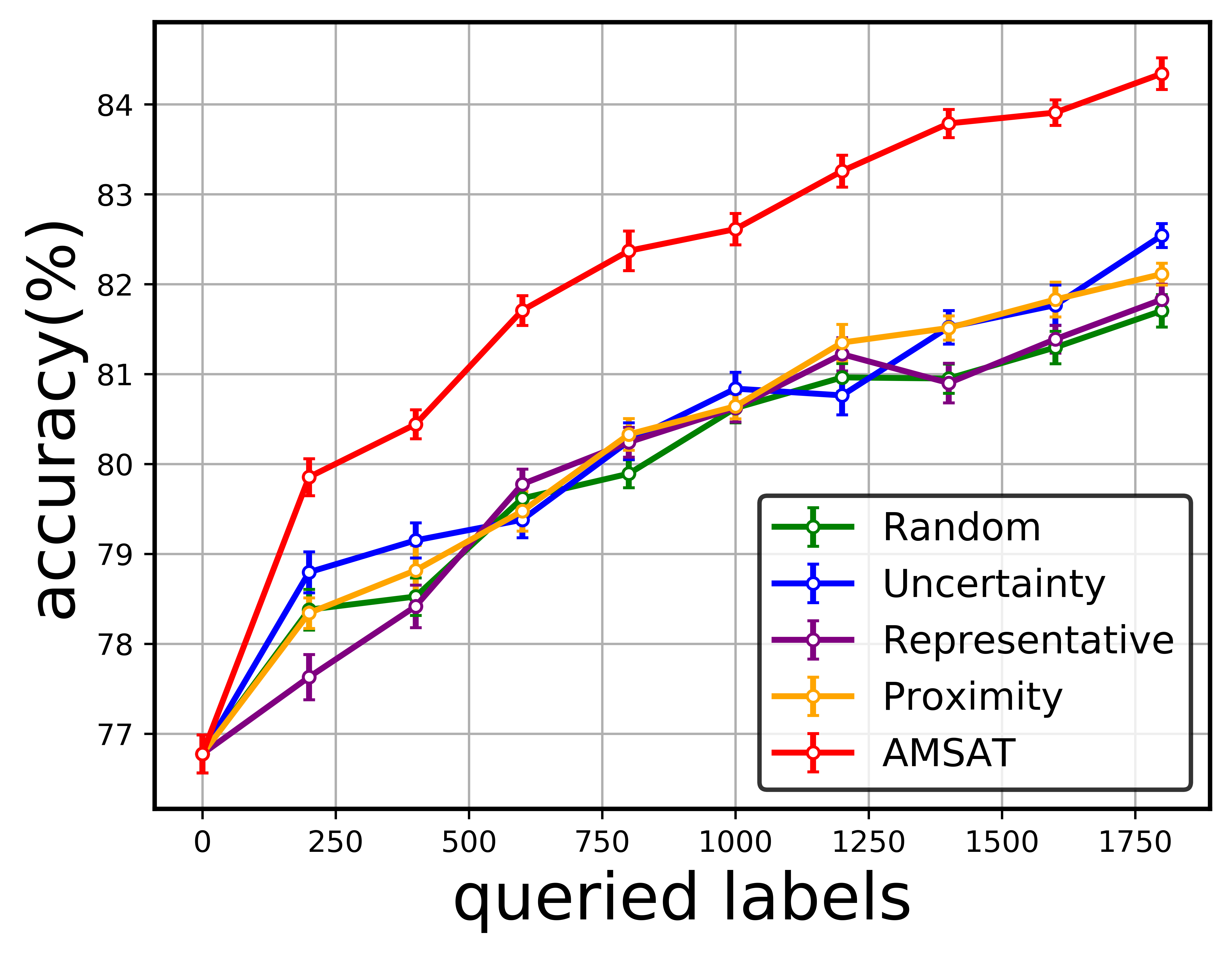} \label{fig:2a} 
  } 
  \hfill 
  \subfigure[Dvd]{% 
    \includegraphics[width=.23\columnwidth, height=.23\columnwidth]{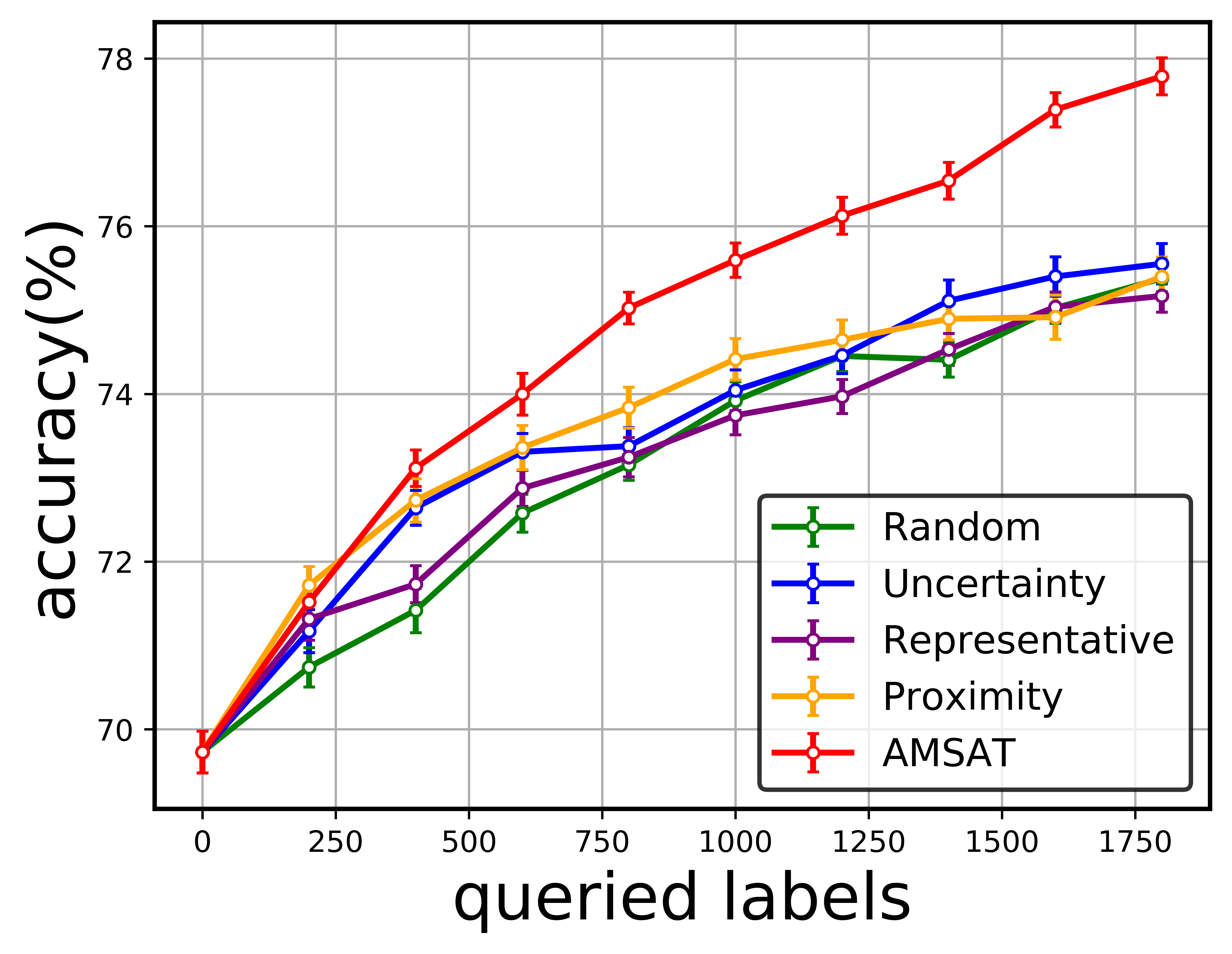} \label{fig:2b} 
  }
  \hfill 
  \subfigure[Sports]{% 
    \includegraphics[width=.23\columnwidth, height=.23\columnwidth]{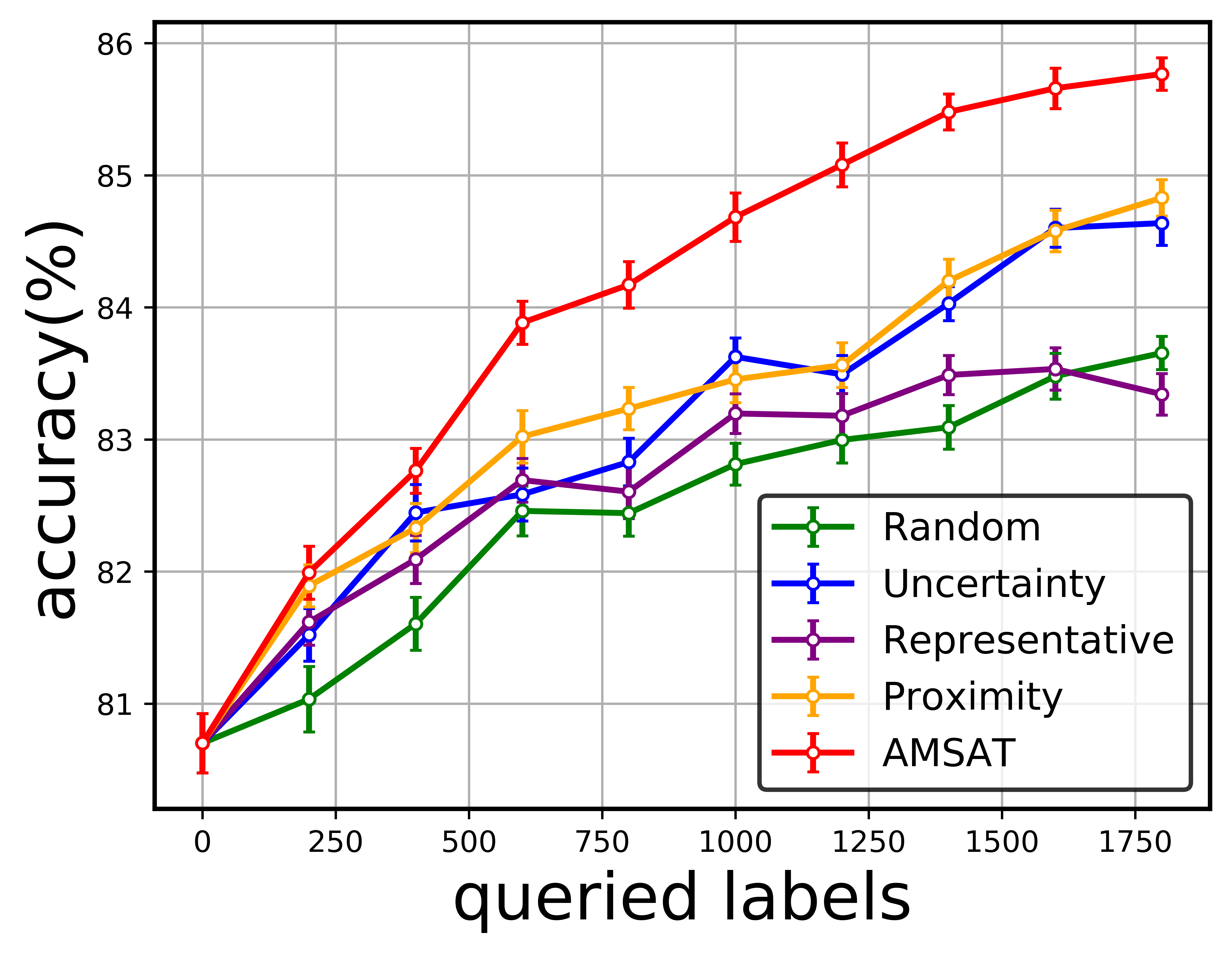} \label{fig:2c} 
  } 
  \hfill 
  \subfigure[Kitchen]{% 
    \includegraphics[width=.23\columnwidth, height=.23\columnwidth]{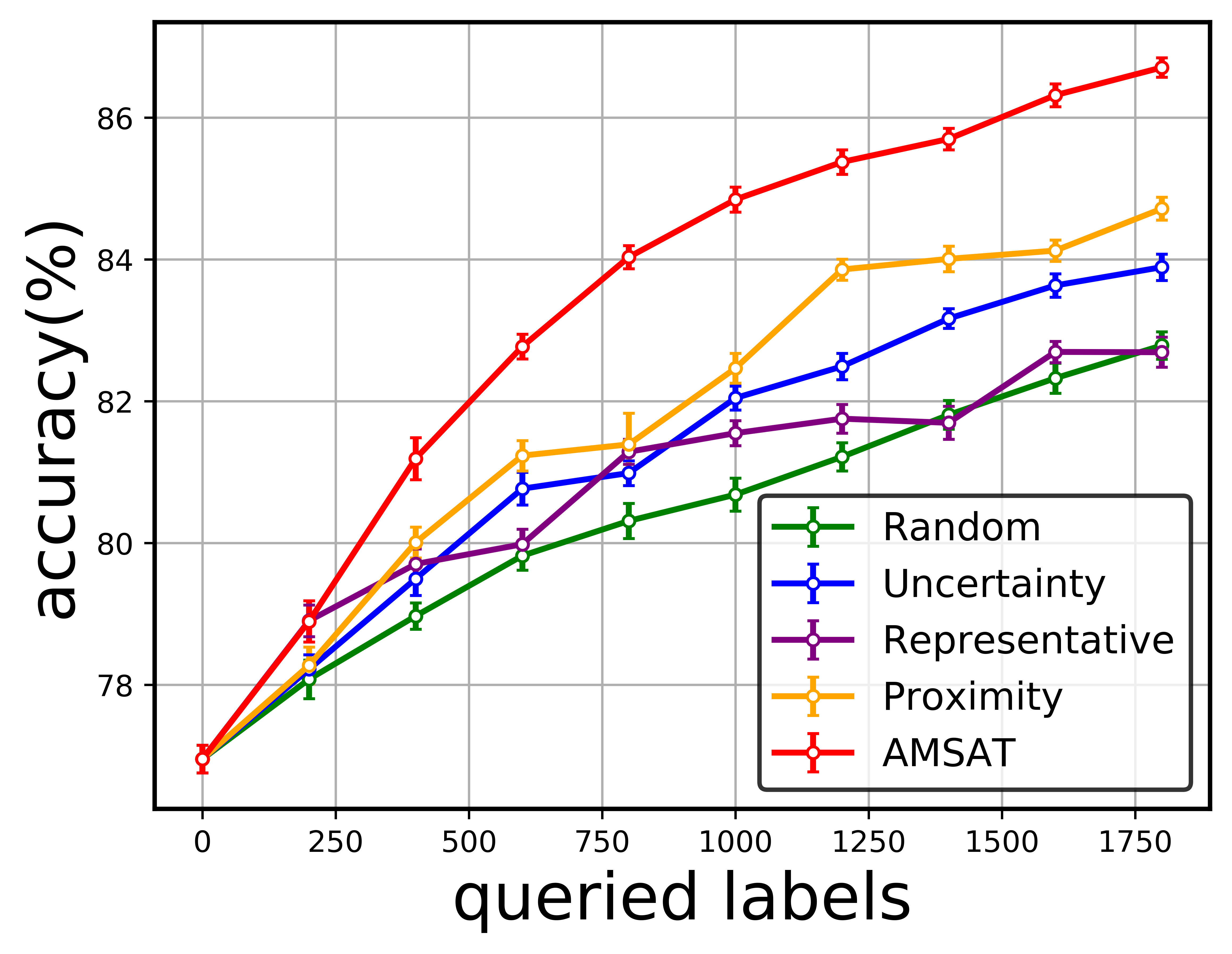} \label{fig:2d} 
  } 
\caption{Performance comparison of active learning methods on \textit{Sentiment}: initial labeled fractions randomly selected from $\{1\%,5\%,15\%,30\%\}$.}
\label{fig:2}
\end{center}
\end{figure*}

\begin{figure*}[tbh]
\begin{center}
    \subfigure[$\{5\%,15\%,30\%\}$]{ 
    \includegraphics[width=.23\columnwidth, height=.23\columnwidth]{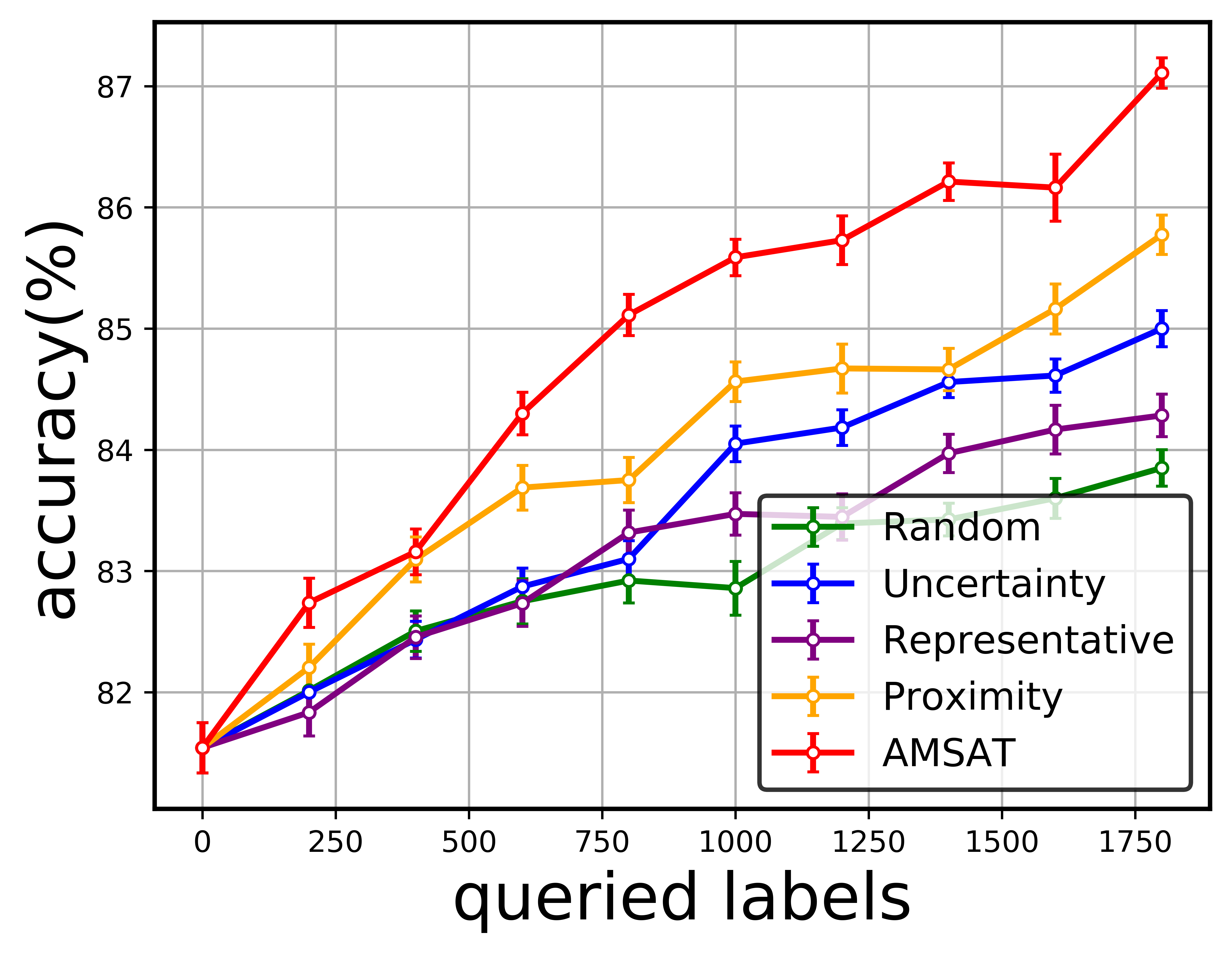} \label{fig:3a} 
  } 
  \hfill 
  \subfigure[$\{1\%,30\%,60\%\}$]{% 
    \includegraphics[width=.23\columnwidth, height=.23\columnwidth]{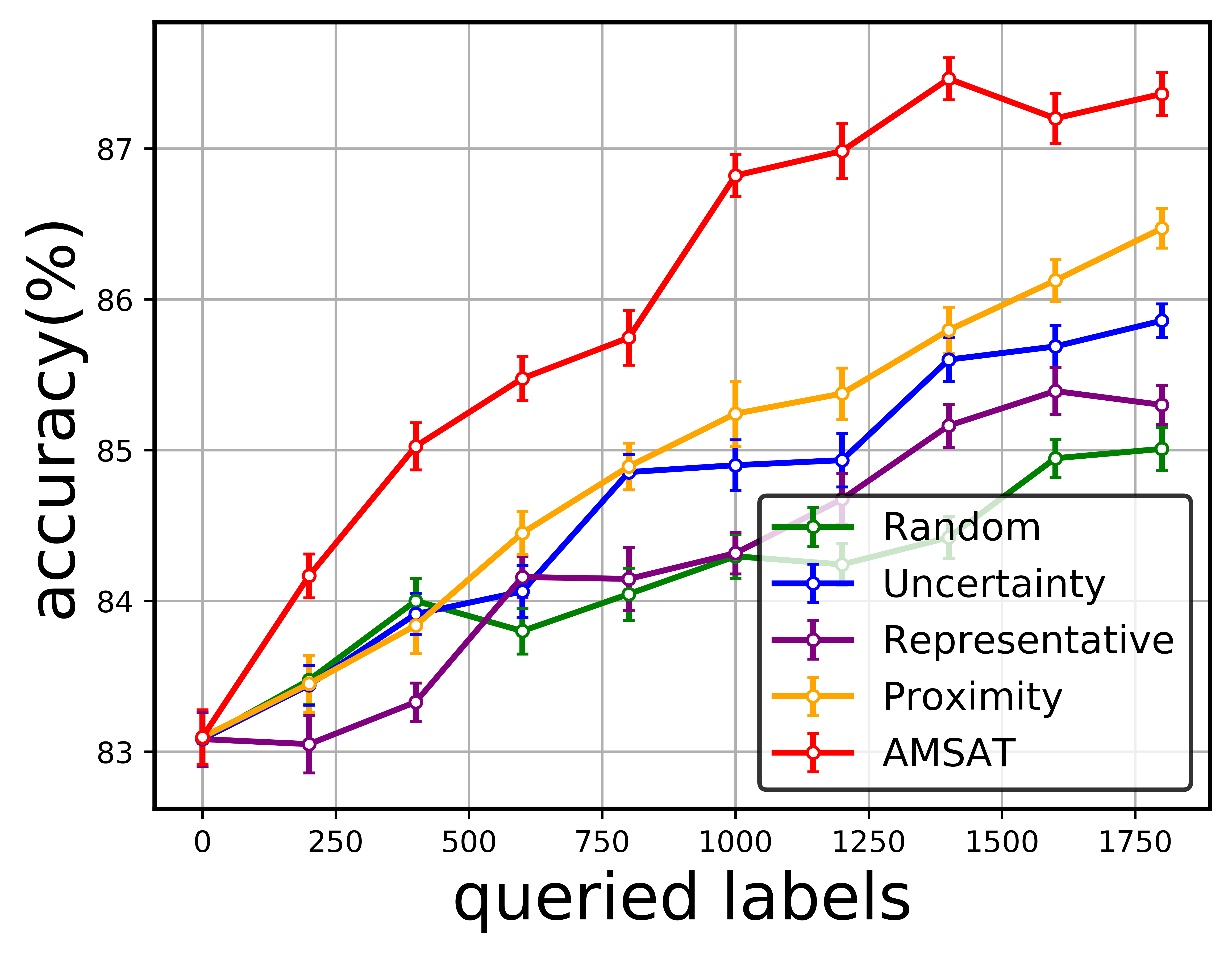} \label{fig:3b} 
  }
  \hfill 
  \subfigure[Electronics]{ 
    \includegraphics[width=.23\columnwidth, height=.23\columnwidth]{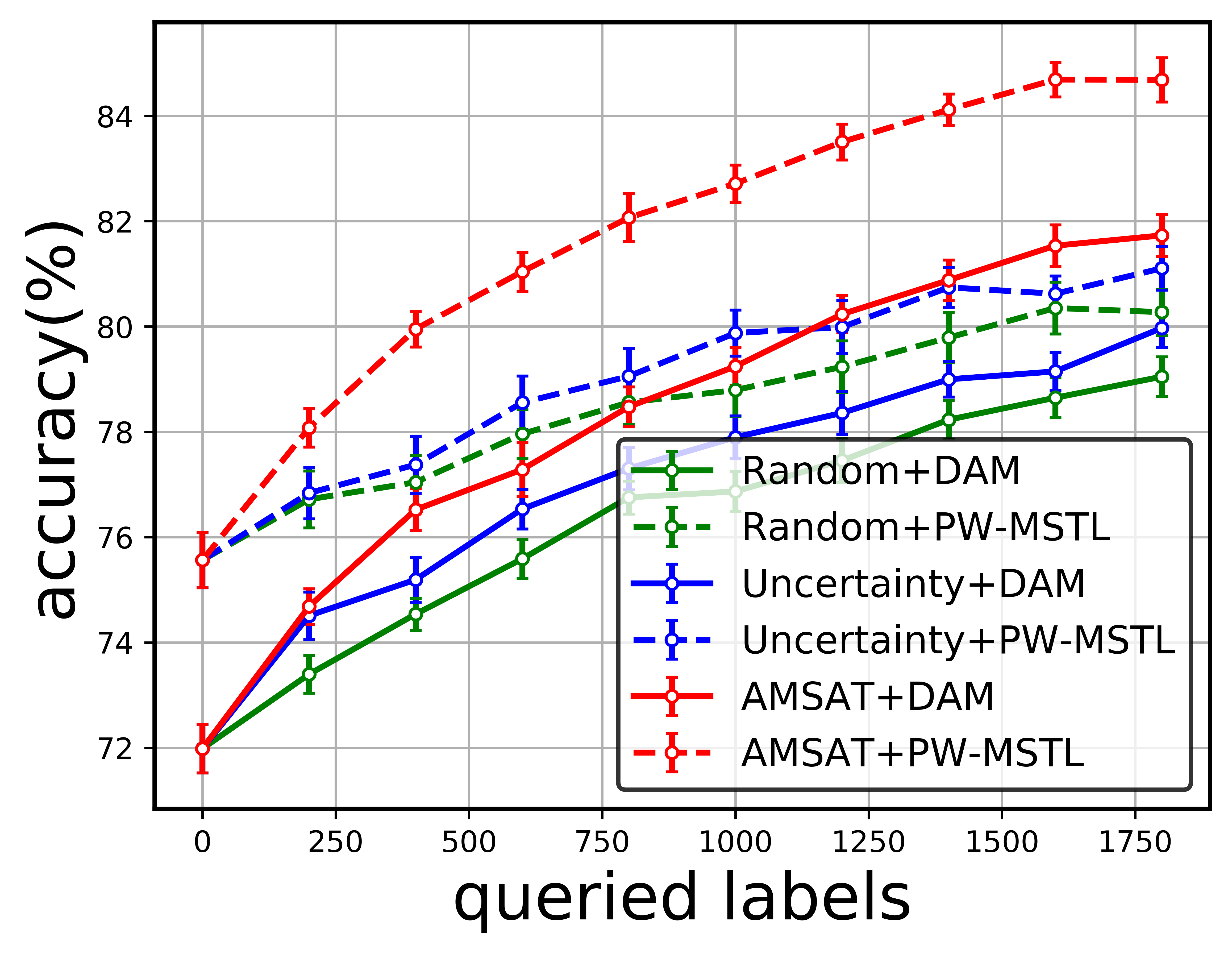} \label{fig:3c} 
  } 
  \hfill
  \subfigure[Book]{% 
    \includegraphics[width=.23\columnwidth, height=.23\columnwidth]{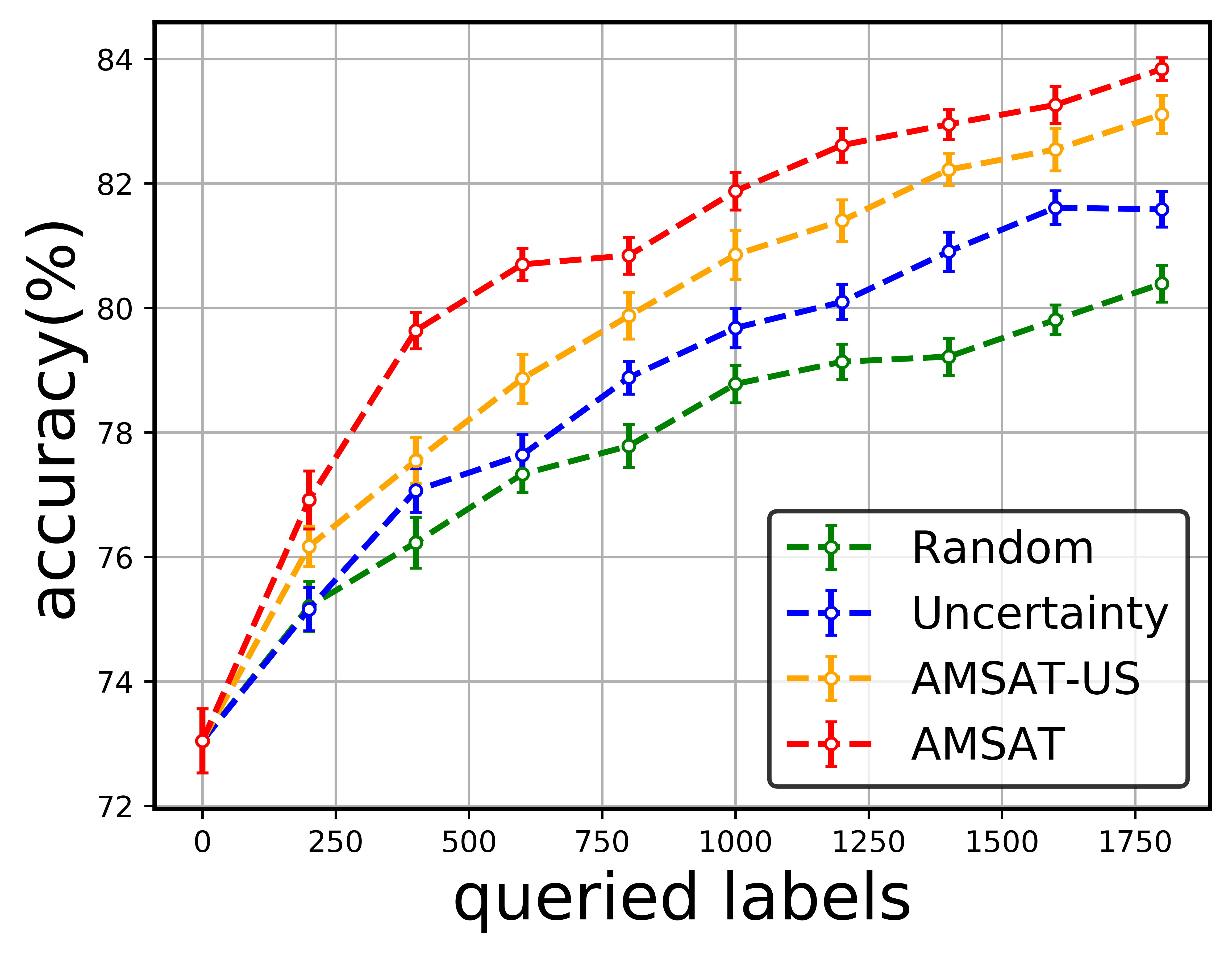} \label{fig:3d} 
  }
\caption{Performance comparison of different labeled data initialization for Kitchen in (a) and (b);  (c) Performance comparison of different combinations of transfer learning methods and active learning methods; (d) Evaluation of the proposed source picking and example picking strategies by comparing AMSAT against Uncertainty and AMSAT-US.}
\label{fig:3}
\end{center}
\end{figure*}

\textbf{Results and Discussion.} The performance curve evaluated using PW-MSTL with increasing queries are plotted in Figure \ref{fig:2}. Due to space limitation, only a small fraction of results are presented and we only show results on \textit{Sentiment} here because other datasets are less interesting due to  diminishing gains. However, it should be noted that our proposed AMSAT method outperforms other methods with $p<0.0001$ significance in a two-tailed test in all datasets. Figure \ref{fig:2} shows results with initial source labeled fractions randomly chosen from $\{1\%,5\%,15\%,30\%\}$, same as experiments in Table \ref{table1}. We can observe that AMSAT consistently outperforms baselines at almost all stages. The only exception is that Proximity sampling performs better at the early stage when there exists a source domain that is particularly close to the target domain, as in the case of \textbf{dvd} in Figure \ref{fig:2b}. Figure \ref{fig:3a} and \ref{fig:3b} show examples of method performance using a different data initialization on the \textbf{kitchen} domain. While they show similar patterns compared to Figure \ref{fig:2d} (AMSAT consistently outperforming other baselines), we observe that AMSAT has comparable ending points in two plots due to diminishing gains. This suggests that performing active learning when sources have very uneven reliabilities is hard.

For the baselines, we observe that they are quite unstable when evaluated using PW-MSTL. For example, in some domains Uncertainty sampling performs better than Random sampling as in Figure \ref{fig:2d} while it is not the case in other domains such as in Figure \ref{fig:2a}. When evaluated using other transfer learning methods such as DAM, often other baselines outperform Random sampling as expected. To show that combining PW-MSTL and AMSAT is superior, we compare several combinations of transfer learning and active learning methods. A typical example is shown in Figure \ref{fig:3c}. Notice that the curves with the same active learning method are evaluated using different transfer learning methods on the same queried data. We observe that while AMSAT still performs better than baselines when evaluated using DAM, the performance gap is larger when evaluated using PW-MSTL. The combination of PW-MSTL and AMSAT significantly outperforms the rest, showing a synergy between the proposed methods.

Finally, we evaluate the effectiveness of both picking the source as line 14 in Algorithm \ref{alg:2} and picking the example according to Eq.(\ref{eq:8}). We add a baseline AMSAT-US which performs exactly the same as Algorithm \ref{alg:2} except that it picks the example according to uncertainty sampling in line 15. Therefore, the added baseline utilizes the proposed source picking strategy but not the example picking strategy. Figure \ref{fig:3d} showcases the comparison among the four methods. We can observe that: (i) AMSAT-US performs better than Uncertainty, indicating the effectiveness of source picking strategy proposed; (ii) AMSAT outperforms AMSAT-US, showing the effectiveness of example picking strategy.

\section{Conclusion}

We study a new research problem of transfer learning with multiple sources exhibiting reliability divergences, and explore two related tasks. The contributions of this paper are: (1) we propose a novel peer-weighted multi-source transfer learning (PW-MSTL) method that makes robust predictions in the described scenario, (2) we study the problem of active learning on source domains and propose an adaptive multi-source active transfer (AMSAT) framework to improve source reliability and performance in the target domain, and (3) we demonstrate the efficacy of utilizing inter-source relationships in the multi-source transfer learning problem. Experiments on both synthetic and real world datasets demonstrated the effectiveness of our methods.

\end{document}